\documentclass[runningheads]{llncs}
\usepackage[T1]{fontenc}
\usepackage{graphicx}
\usepackage{adjustbox}
\usepackage{cite}
\usepackage{multirow}
\usepackage{tikz}
\usetikzlibrary{fit}
\usetikzlibrary{shapes,arrows}
\usepackage{siunitx}
\usepackage{breqn}
\usepackage{lettrine}
\usepackage{etoolbox}
\usepackage{algorithm}
\usepackage{orcidlink}
\usepackage{hyperref}
\usepackage{makecell} 

\usepackage[noend]{algpseudocode}
\begin{document}
\title{Tool-to-Agent Retrieval: Bridging Tools and Agents for Scalable LLM Multi-Agent Systems}
\titlerunning{Tool-to-Agent Retrieval}
%
\author{Elias Lumer \and
Faheem Nizar \and
Anmol Gulati \and \\
Pradeep Honaganahalli Basavaraju \and
Vamse Kumar Subbiah}
\authorrunning{E. Lumer et al.}
\institute{PricewaterhouseCoopers U.S.A.}

%
%
\maketitle              
\begin{abstract}
Recent advances in LLM Multi-Agent Systems enable scalable orchestration of sub-agents, each coordinating hundreds or thousands of tools or Model Context Protocol (MCP) servers. However, existing retrieval methods typically match queries against coarse agent-level descriptions before routing, which obscures fine-grained tool functionality and often results in suboptimal agent selection. We introduce Tool-to-Agent Retrieval, a unified framework that embeds both tools and their parent agents in a shared vector space and connects them through metadata relationships. By explicitly representing tool capabilities and traversing metadata to the agent level, Tool-to-Agent Retrieval enables granular tool-level or agent-level retrieval, ensuring that agents and their underlying tools or MCP servers are equally represented without the context dilution that arises from chunking many tools together. Evaluating Tool-to-Agent Retrieval across eight embedding models, our approach achieves consistent improvements of 19.4\% in Recall@5 and 17.7\% in nDCG@5 over previous state-of-the-art agent retrievers on the LiveMCPBench benchmark.
\end{abstract}
 
\keywords{Large Language Models \and Tool Retrieval \and Agent Routing \and Multi-Agent Systems \and Model Context Protocol (MCP)}

\section{Introduction} Recent advancements in Large Language Model (LLM) agents and the Model Context Protocol (MCP) enable assistants to discover, equip, and use large collections of external tools and MCP servers at inference time \cite{fei2025mcpzeroactivetooldiscovery, lumer2025memtooloptimizingshorttermmemory}. In practice, a single assistant may delegate to specialized sub-agents for code analysis, databases, or web search, each bundling dozens of tools behind a single interface \cite{du_anytool_2024}. A central challenge is routing: given a user query, should the system select a specific tool or leverage an entire agent (e.g., an MCP server) that provides a coherent set of tools? Forwarding all tool descriptions to the model is impractical, e.g., one MCP server with 26 tools can consume over 4{,}600 tokens, making efficient retrieval essential for scalability \cite{yuan2024easytool_2024,wu2024sealtoolsselfinstructtoollearning}.

Existing strategies typically fall into two camps. Agent-first pipelines match the query against a brief agent description, then operate only within that agent, which can hide highly relevant tools whose parent description does not obviously align with the query. Conversely, tool-only retrieval treats each tool independently and ignores the complementary benefits of the surrounding bundle on multi-step tasks \cite{lumer2024toolshedscaletoolequippedagents}. Recent benchmarks have highlighted these challenges in multi-step tool selection, evaluating agents across diverse tool repositories \cite{li_api_bank_2023,qin2023toolllmfacilitatinglargelanguage,huang2024metatoolbenchmarklargelanguage}. In practice, single-tool matches may underperform when a workflow benefits from a coordinated set of tools, while agent-first routing may overlook fine-grained capabilities. What is needed is a retrieval mechanism that can operate at both levels, return a focused tool when the query is specific, or return an entire agent when broader capability is advantageous, without committing to a brittle two-stage pipeline.

\begin{figure}[t]
    \centering
    \includegraphics[width=\linewidth]{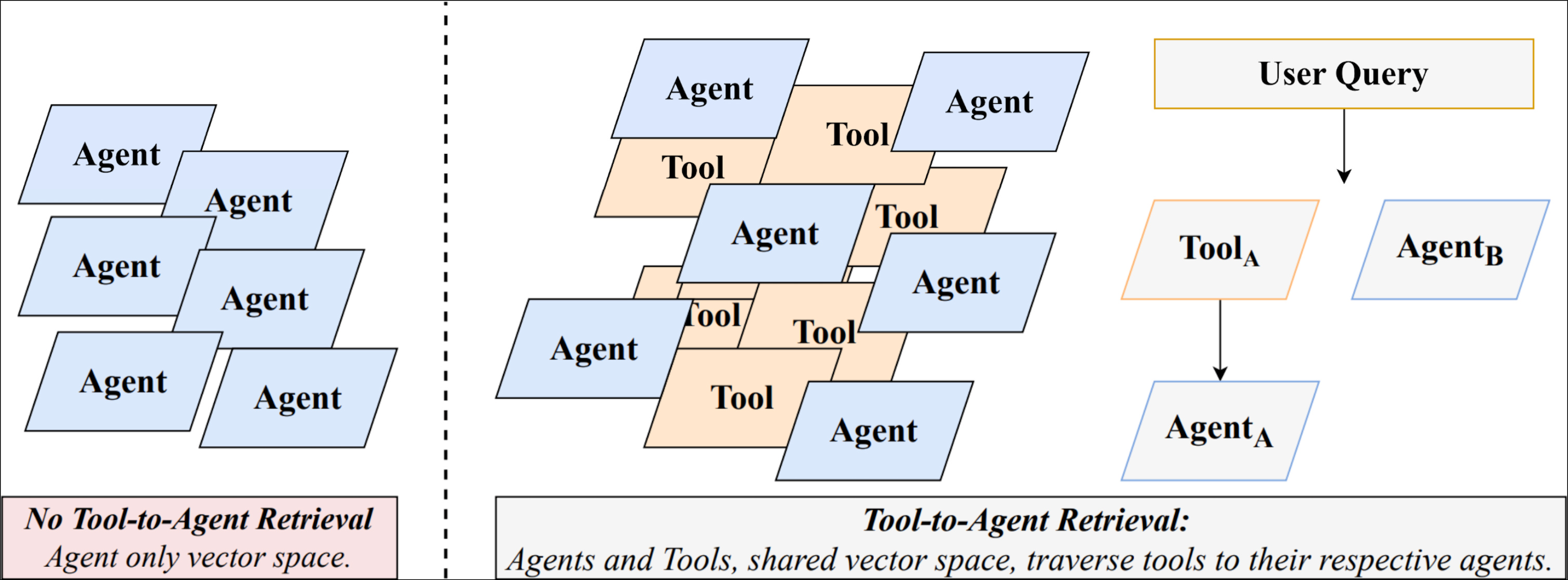}
    \caption{Comparison of agent-only retrieval (left) and the proposed Tool-to-Agent Retrieval (right), which embeds tools and agents in a shared vector space to support joint retrieval and traversal.}
    \label{fig:main_diagram}
\end{figure}

We introduce \textit{Tool-to-Agent Retrieval}, which represents tools and their parent agents in the same vector space and links them through explicit metadata (Figure~\ref{fig:main_diagram}). At query time, retrieval runs over a joint index and returns either an individual tool or an agent bundle, depending on which better fits the query. By modeling fine-grained tool semantics while retaining agent context via tool$\rightarrow$agent links (and traversing those links at retrieval time), the method avoids the context dilution that arises when many tools are collapsed into a single coarse description. This enables one-pass decisions about what to equip (a tool vs.\ an agent), improving routing for both focused and multi-step queries.

\textbf{\textit{Related Work.}} A growing line of work builds tool retrieval for LLM agents by embedding tool descriptions and metadata in a tool knowledge base and selecting top-$k$ tools per query \cite{lumer2025scalemcpdynamicautosynchronizingmodel,lumer2024toolshedscaletoolequippedagents, lumer2025graphragtoolfusion, yuan2024craftcustomizingllmscreating}. Retrieval-augmented generation (RAG) enables LLM agents to access external knowledge and tools through semantic search \cite{gao2024retrievalaugmentedgenerationlargelanguage}, with recent hybrid approaches combining knowledge graphs and vector retrieval to improve tool selection accuracy \cite{sarmah2024hybridragintegratingknowledgegraphs}. Systems such as PLUTO and Re-Invoke refine dense retrieval for tool selection and demonstrate strong single-pass performance at scale \cite{huang2024planningeditingretrieveenhanced,chen2024reinvoketoolinvocationrewriting}. Graph-based approaches have emerged to structure tool relationships, including ToolNet and ControlLLM which organize tools through graph structures to enable more sophisticated navigation \cite{liu2024toolnetconnectinglargelanguage,liu2023controlllmaugmentlanguagemodels}. Iterative methods like ToolReAGT decompose complex tasks and retrieve tools step-by-step, improving recall on multi-step problems \cite{braunschweiler2025toolreagt, anantha2023protipprogressivetoolretrieval}. Multi-hop reasoning frameworks further enhance retrieval by chaining intermediate steps \cite{trivedi_interleaving_2023,yao_react_2023}. In parallel, hierarchical pipelines first choose the MCP server and then a tool inside it to control prompt cost, but this agent-first routing can suppress fine-grained matches \cite{fei2025mcpzeroactivetooldiscovery, zheng2024toolrerankadaptivehierarchyawarereranking}. Our approach differs by jointly representing tools and agents and retrieving over a single, linked space, so the system does not have to commit upfront to tool-only or agent-first selection.

\textbf{\textit{Contributions.}} We make the the following contributions through this work:
\begin{enumerate}
    \item \textbf{Unified Retrieval Framework:} We introduce a novel tool retrieval strategy that embeds both tools and their parent agents in a shared vector space, linked through tool-to-agent metadata traversal, enabling unified retrieval and achieving state-of-the-art performance.
    \item \textbf{Granular Routing Mechanism:} We propose a retrieval procedure that preserves fine-grained tool-level detail while retaining agent-level context, mitigating context dilution from coarse summaries and improving robustness on multi-step queries.
    \item \textbf{Comprehensive Evaluation:} We evaluate the method on LiveMCPBench across eight embedding models, demonstrating 17.7\% (recall@5) and 19.4\% (nDCG@5) improvements over previous state-of-the-art approaches. 
\end{enumerate}

\begin{algorithm}[t]
\footnotesize
\caption{\small Combined Tool--Agent Top-$K$ Retrieval (Tool-to-Agent Retrieval)}
\label{alg:cta-select}
\begin{algorithmic}[1]
\State \textbf{Input}: query $q$, corpus $\mathcal{C}$ (agents $\cup$ tools), type $\tau(\cdot)\in\{\text{agent},\text{tool}\}$, owner map $\mathrm{own}(\cdot)$, similarity $s(q,\cdot)$, cutoffs $N$, $K$
\State \textbf{Output}: set of agents $\mathcal{A}^\star$ with $|\mathcal{A}^\star|= K$
\State $\mathcal{L} \gets \textsc{TopN}(q,\mathcal{C},N)$ \Comment{rank by $s$ and return ordered $[e^{(1)},\ldots,e^{(N)}]$}
\State $\mathcal{A} \gets \emptyset$; \ $i \gets 1$
\While{$|\mathcal{A}| < K$ \textbf{ and } $i \le N$}
  \State $e \gets \mathcal{L}[i]$
  \If{$\tau(e)=\text{agent}$}
     \State $a \gets e$
  \ElsIf{$\tau(e)=\text{tool}$ \textbf{ and } $\mathrm{own}(e)$ \textbf{is defined}}
     \State $a \gets \mathrm{own}(e)$
  \Else
     \State $i \gets i+1$; \textbf{continue} \Comment{skip if owner missing/undefined}
  \EndIf
  \If{$a \notin \mathcal{A}$}
     \State $\mathcal{A} \gets \mathcal{A} \cup \{a\}$
  \EndIf
  \State $i \gets i+1$
\EndWhile
\State \textbf{return} $\mathcal{A}^\star \gets \mathcal{A}$
\end{algorithmic}
\end{algorithm}

\section{Approach: Tool-to-Agent Retrieval}

As illustrated in Figure~\ref{fig:main_diagram}, \textit{Tool-to-Agent Retrieval} embeds both tools and their parent agents in a unified vector space, explicitly linking each tool to its parent agent through metadata relationships. We consider a catalog of MCP servers with their corresponding \emph{agents}, denoted as \(a \in \mathcal{A}\). Each agent \(a\) owns a set of \emph{tools} \(\mathcal{T}_a\), consisting of API calls, functions, or actions exposed by the agent. The combined system is modeled as a bipartite graph \(G = (\mathcal{A}, \mathcal{T}, E)\), where edges \(E\) represent ownership relations between tools and agents. Graph-based modeling of tool relationships enables structured navigation and reasoning over tool capabilities \cite{sun2024thinkongraphdeepresponsiblereasoning,jin2024graphchainofthoughtaugmentinglarge}. This unified representation directly addresses the limitations of single-level retrieval. Retrieving solely by agent descriptions can obscure fine-grained functional capabilities at the tool level, while retrieving tools independently discards important execution context such as authentication, parameter inference, or access policies maintained at the agent level. By integrating both levels, the retriever can surface relevant tools without losing their surrounding agent context. Furthermore, every query or sub-query ultimately resolves to an executable agent, ensuring that the retrieved entity can act upon the user request.


\paragraph{\textbf{Indexing.}} 
We construct a unified \emph{tool--agent catalog} \(\mathcal{C}\) that integrates both tools and agents for retrieval. The catalog is composed of two corpora: the \emph{tool corpus} \(\mathcal{C_T}\) and the \emph{agent corpus} \(\mathcal{C_A}\).

The Tool Corpus \(\mathcal{C_T} \subset \mathcal{C}\) contains tool names and descriptions that are directly indexed for retrieval. Each tool entry includes metadata explicitly linking it to its parent MCP server or agent, denoted as \(owner(\mathcal{T}) = \mathcal{A}\). This mapping enables traversal from a retrieved tool to the corresponding executable agent during query resolution, drawing on graph-based representation techniques for knowledge integration \cite{Pan_2024,peng2024graphretrievalaugmentedgenerationsurvey}.

The Agent Corpus \(\mathcal{C_A} \subset \mathcal{C}\) similarly contains agent names and descriptions, representing higher-level capabilities and serving as parent nodes within the retrieval graph.



\paragraph{\textbf{Retrieval.}} 
The retrieval process modifies the standard top-\(K\) ranking procedure. The objective is to identify the top-\(K\) most relevant agents for a given query or sub-query. To achieve this, we first retrieve the top \(N \gg K\) entities from the unified tool--agent catalog \(\mathcal{C}\), ranked by semantic similarity to the query. This approach combines semantic and lexical matching strategies to improve recall \cite{kuzi2020leveragingsemanticlexicalmatching}, utilizing BM25 \cite{robertson_probabilistic_2009} alongside dense vector retrieval. The corresponding parent agents are then aggregated, and the top-\(K\) unique agents are selected. The complete retrieval procedure is detailed in Algorithm~\ref{alg:cta-select}.


\paragraph{\textbf{Query.}} 
The input to the Tool-to-Agent Retriever can be the original user query, the decomposed sub-steps derived from it, or a combination of both. We evaluate two query paradigms. 

The first, Direct Querying, uses the user's high-level question directly as the retrieval query, without any pre-processing. This approach retrieves the top-\(K\) most relevant agents or tools for the overall task. 

The second, Step-wise Querying, decomposes the original query into a sequence of smaller sub-tasks. Each step is then submitted independently to the retriever, allowing the system to identify different agents as needed across multi-step workflows. This decomposition strategy aligns with reasoning-based retrieval planning methods that break complex queries into manageable sub-goals \cite{joshi_reaper_2024,khattab_demonstrate-search-predict_2023}. Query rewriting techniques further enhance retrieval by reformulating queries for improved semantic matching \cite{ma_query_2023}. This step-wise procedure is the primary setting used in our evaluations.

\begin{table}[t]
\caption{Results on the LiveMCPBench benchmark comparing Tool-to-Agent Retrieval with baselines (BM25, Q.Retrieval, ScaleMCP, and MCPZero). 
Metrics are Recall@K, mAP@K, and nDCG@K for $K \in \{1, 5, 10\}$.}
\label{tab:results_primary}
\centering
\begingroup
\small 
\setlength{\tabcolsep}{3pt} 
\renewcommand{\arraystretch}{1.05} 
\begin{tabular}{l|ccc|ccc|ccc}
\hline
\textbf{Approach} & \multicolumn{3}{c|}{\textbf{Recall}} & \multicolumn{3}{c|}{\textbf{mAP}} & \multicolumn{3}{c}{\textbf{nDCG}} \\
 & @1 & @3 & @5 & @1 & @3 & @5 & @1 & @3 & @5 \\
\hline
BM25 & 0.20 & 0.20 & 0.20 & 0.12 & 0.12 & 0.12 & 0.14 & 0.14 & 0.14 \\
Q.Retrieval & 0.31 & 0.47 & 0.56 & 0.31 & 0.31 & 0.24 & 0.31 & 0.35 & 0.32 \\
MCPZero & 0.44 & 0.66 & 0.70 & 0.45 & 0.39 & \underline{0.31} & 0.45 & 0.46 & \underline{0.41} \\
ScaleMCP & \underline{0.49} & \underline{0.68} & \underline{0.74} & \underline{0.49} & \underline{0.40} & 0.29 & \underline{0.49} & \underline{0.48} & 0.40 \\
\textbf{Tool-to-Agent Retrieval} & \textbf{0.61} & \textbf{0.77} & \textbf{0.83} & \textbf{0.61} & \textbf{0.49} & \textbf{0.34} & \textbf{0.61} & \textbf{0.56} & \textbf{0.46} \\
\hline
\end{tabular}
\endgroup
\end{table}

\section{Experiments and Evaluation}

We evaluate the effectiveness of the proposed Tool-to-Agent Retriever in comparison to agent-only retrieval methods. Our primary hypothesis is that a unified retriever operating over both tools and agents will outperform approaches that retrieve agents alone. In the MCP context, particularly under hierarchical querying \cite{fei2025mcpzeroactivetooldiscovery,lumer2025scalemcpdynamicautosynchronizingmodel}, this improvement extends to the initial stage of routing, where identifying the correct MCP server enables access to its associated tools for cross-connected operations. Recent benchmarking efforts have highlighted the challenges of stateful, conversational tool use across multi-turn interactions \cite{lu2024toolsandboxstatefulconversationalinteractive,zhong2025complexfuncbenchexploringmultistepconstrained}.

The experimental setup evaluates retrieval accuracy in identifying the correct agents or MCP servers for each query. We report standard information retrieval metrics, including Recall, mean average precision (mAP), and normalized discounted cumulative gain (nDCG), to quantify the performance improvements of the proposed approach.


\paragraph{\textbf{Dataset and Evaluation Protocol.}}
We evaluate our approach on the LiveMCPBench dataset \cite{mo2025livemcpbenchagentsnavigateocean,LiveMCPBench_eval_questions_github}, which includes 70 MCP servers and 527 tools, along with 95 real-world questions annotated with step-by-step breakdowns and relevant tool–agent mappings. This structure enables fine-grained, step-level evaluation of retrieval performance. On average, each question spans 2.68 steps and involves 2.82 tools and 1.40 MCP agents.

To isolate the contribution of tool-level information, we also construct an \emph{agent-only} baseline dataset containing only MCP server names and descriptions. During inference, we employ step-wise querying and compare Tool-to-Agent Retrieval against BM25 \cite{INR-019}, ScaleMCP \cite{lumer2025scalemcpdynamicautosynchronizingmodel}, and MCPZero \cite{fei2025mcpzeroactivetooldiscovery}.


\paragraph{\textbf{Retriever Setup.}} 
We evaluate retrieval performance across several embedding models. Specifically, we use 8 embedding models, both closed and open source \cite{vertex_embedding_models, titan_embedding_models, openai_embedding_models}. The datasets are embedded using each model, and semantic similarity search is performed to retrieve relevant entities. We first retrieve the top \(N \gg K\) entities from the tool--agent catalog and then select the top-\(K\) unique agents using Algorithm~\ref{alg:cta-select}. Retrieval accuracy is computed by comparing the retrieved agents against the ground-truth agents associated with each query in the evaluation set.


\paragraph{\textbf{Results.}} 
Tables~\ref{tab:results_primary} and~\ref{tab:retriever_comparison} show that Tool-to-Agent Retrieval consistently outperforms prior methods across Recall, mAP, and nDCG metrics. Our approach achieves superior performance over all baselines, with gains observed across multiple embedding families including Vertex AI, Gemini, Titan, OpenAI, and MiniLM. 

These improvements arise primarily from the richer retrieval corpus that jointly indexes both tools and agents, enabling finer-grained semantic alignment. Importantly, the performance lift is not attributable to tool-level retrieval alone. The joint index supports nuanced matches while preserving agent context, as evidenced by 39.13\% of retrieved top-\(K\) items originating from the agent corpus \(\mathcal{C_A}\) and 34.44\% of matched top-\(K\) tools also tracing back to \(\mathcal{C_A}\). Together, these results demonstrate that explicitly linking tools to their parent agents mitigates context dilution and improves multi-step routing without sacrificing fine-grained precision. Across all eight embedding models, Tool-to-Agent Retrieval exhibited remarkably stable improvements, with standard deviations of 0.02 in Recall@5 and 0.01 in nDCG@5 relative to MCPZero. This consistency indicates that the gains are architecture-agnostic and primarily driven by the unified indexing design rather than embedding-specific behavior. The strongest relative improvement was observed on Amazon Titan v2 (Recall@5 improved from 0.66 to 0.85, a +28\% relative gain), while even the compact All-MiniLM-L6-v2 model achieved a +13\% improvement, confirming generalizability across both proprietary and open-source embeddings.

\begin{table}[t]
\centering
\caption{Per-embedding comparison against MCPZero}
\label{tab:retriever_comparison}
\begingroup
\small
\setlength{\tabcolsep}{4pt}
\renewcommand{\arraystretch}{1.05}
\begin{tabular}{l|cc|cc|cc}
\hline
\textbf{Retriever Model} & 
\multicolumn{2}{c|}{\textbf{Recall@5}} & 
\multicolumn{2}{c|}{\textbf{nDCG@5}} & 
\multicolumn{2}{c}{\textbf{mAP@5}} \\ 
\cline{2-7}
 & Ours & MCPZero & Ours & MCPZero & Ours & MCPZero \\
\hline
\footnotesize{Vertex AI text embedding 005} & \textbf{0.87} & \underline{0.74} & \textbf{0.48} & \underline{0.42} & \textbf{0.36} & \underline{0.32} \\
\footnotesize{Gemini Embedding 001} &           \textbf{0.86} & \underline{0.74} & \textbf{0.49} & \underline{0.44} & \textbf{0.37} & \underline{0.34} \\
\footnotesize{Amazon titan embed text v2} &     \textbf{0.85} & \underline{0.66} & \textbf{0.47} & \underline{0.37} & \textbf{0.35} & \underline{0.28} \\
\footnotesize{Amazon titan embed text v1} &     \textbf{0.85} & \underline{0.65} & \textbf{0.48} & \underline{0.39} & \textbf{0.36} & \underline{0.30} \\
\footnotesize{OpenAI text embedding ada 002} &  \textbf{0.83} & \underline{0.70} & \textbf{0.50} & \underline{0.40} & \textbf{0.39} & \underline{0.30} \\
\footnotesize{OpenAI text embedding 3 small} &  \textbf{0.87} & \underline{0.72} & \textbf{0.49} & \underline{0.41} & \textbf{0.36} & \underline{0.31} \\
\footnotesize{OpenAI text embedding 3 large} &  \textbf{0.87} & \underline{0.74} & \textbf{0.50} & \underline{0.42} & \textbf{0.38} & \underline{0.32} \\
\footnotesize{All MiniLM L6 v2} &               \textbf{0.80} & \underline{0.67} & \textbf{0.45} & \underline{0.39} & \textbf{0.33} & \underline{0.29} \\
\hline
\end{tabular}
\endgroup
\end{table}

\section{Conclusion}
Recent advances in LLM multi-agent systems enable scalable orchestration of sub-agents coordinating hundreds or thousands of tools or Model Context Protocol (MCP) servers. We introduce Tool-to-Agent Retrieval, a unified framework for large language model (LLM) multi-agent systems that embeds both tools and their parent agents in a shared vector space, linked through metadata relationships. By explicitly modeling tool capabilities and enabling traversal between tool- and agent-level representations, our approach supports granular retrieval decisions that preserve fine-grained context without the dilution introduced by coarse agent summaries. Experiments across eight embedding models on the LiveMCPBench benchmark demonstrate substantial improvements over prior agent retrievers, with gains of 19.4\% in Recall@5 and 17.7\% in nDCG@5. These results highlight a promising direction for unifying tool and agent selection, motivating future research into retrieval architectures that scale across increasingly complex agent networks.

\clearpage

%
%
%
%
\bibliographystyle{splncs04}
\bibliography{references}
\end{document}